\documentclass[letterpaper]{article} % DO NOT CHANGE THIS
\usepackage{aaai2026}  % DO NOT CHANGE THIS
\usepackage{times}  % DO NOT CHANGE THIS
\usepackage{helvet}  % DO NOT CHANGE THIS
\usepackage{courier}  % DO NOT CHANGE THIS
\usepackage[hyphens]{url}  % DO NOT CHANGE THIS
\usepackage{graphicx} % DO NOT CHANGE THIS
\usepackage{natbib}  % DO NOT CHANGE THIS AND DO NOT ADD ANY OPTIONS TO IT
\usepackage{caption} % DO NOT CHANGE THIS AND DO NOT ADD ANY OPTIONS TO IT
\usepackage{amsmath} % Added for align environment
\usepackage{xcolor}

\usepackage{tcolorbox}
\usepackage{placeins}
\usepackage{booktabs}
\usepackage{makecell}

\usepackage{algorithm}
\usepackage{algorithmic}

\usepackage{newfloat}
\usepackage{listings}
\DeclareCaptionStyle{ruled}{labelfont=normalfont,labelsep=colon,strut=off} % DO NOT CHANGE THIS
\floatstyle{ruled}
\newfloat{listing}{tb}{lst}{}
\floatname{listing}{Listing}
\title{Mitigating Self-Preference by Authorship Obfuscation}
\author{
    Taslim Mahbub, Shi Feng\\
}
\affiliations{
    George Washington University\\
    taslim.mahbub,shi.feng@gwu.edu
}

\begin{document}

\maketitle

\begin{abstract}
Language models (LMs) judges are widely used to evaluate the quality of LM outputs.
Despite many advantages, LM judges display concerning biases that can impair their integrity in evaluations. One such bias is self-preference: LM judges preferring their own answers over those produced by other LMs or humans. The bias is hard to eliminate as frontier LM judges can distinguish their own outputs from those of others, even when the evaluation candidates are not labeled with their sources. In this paper, we investigate strategies to mitigate self-preference by reducing the LM judges' ability to recognize their own outputs. We apply black-box perturbations to evaluation candidates in pairwise comparison to obfuscate the authorship and reduce self-recognition. We find that perturbations as simple as synonym replacement for a few words predictably reduce self-preference. However, we also uncover fundamental challenges to eliminating the bias: when we extrapolate our perturbations to a more complete neutralization of stylistic differences between the evaluation candidates, self-preference recovers. Our findings suggest that self-recognition and self-preference can happen on many semantic levels, and complete mitigation remains challenging despite promising initial results.
\end{abstract}

% Uncomment the following to link to your code, datasets, an extended version or similar.
% You must keep this block between (not within) the abstract and the main body of the paper.
\begin{links}
    \link{Code}{https://github.com/Taslim-M/mitigatingselfpreference}
    % \link{Datasets}{https://aaai.org/example/datasets}
    % \link{Extended version}{https://aaai.org/example/extended-version}
\end{links}

\section{Introduction}

\begin{figure}[t]
\centering
\includegraphics[width=0.75\columnwidth]{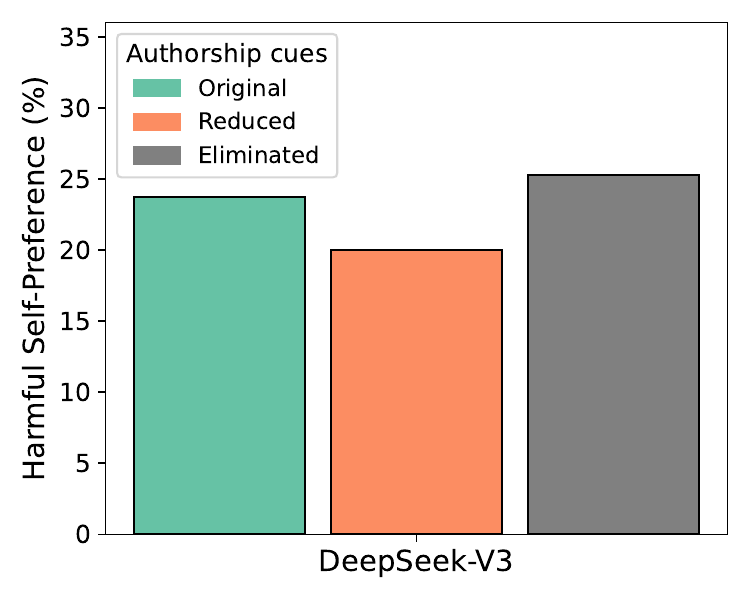} 
\caption{Harmful self-preference of DeepSeek-V3 in pairwise comparison. When we perturb the answer pair to reduce superficial cues (e.g., word choice) that the judge can use to infer model identity, harmful self-preference also reduces. However, the effect is reversed when we eliminate identity cues by paraphrasing.}
\label{fig:headline}
\end{figure}

Language models (LMs) are frequently used in place of human evaluators to judge the quality of LM outputs.
These LM judges are now widely used for benchmarking \cite{zheng2023judging, bai2023benchmarking}, to help train LMs as reward models \cite{leike2018scalable, stiennon2020learning}, and to guide inference-time compute \cite{saunders2022self, chen2023teaching}.
This ``LM-as-a-judge'' approach has clear scalability advantages, but also suffers from biases that can hurt the integrity of evaluations and derail model development.
One particularly concerning bias is self-preference \cite{zheng2023judging, liu2023llms, panickssery2024llm, li2025preference, goel2025greatmodelsthinkalike, ohi2024likelihood, chen2025llm}, where the LM judge prefers its own answers over those produced by other models or humans, even when the alternative's quality is objectively higher.
% This is particularly concerning given how a small number of frontier models are likely used by the majority of benchmarks due to their superior capabilities---the benchmarks might collectively and disproportionately favor models similar to those judges.
Self-preference is particularly harmful when the judge prefers its own incorrect or unsafe answers over correct answers from other models \cite{chen2025llm}.
Judges with harmful self-preference can amplify untruthfulness and deceitfulness, a concern for both capability and safety.

% In the context of AI control, untrusted monitoring often involves using a separate instance of the same model to evaluate a generated action before it is executed. However, suppose the model exhibits bias toward outputs produced by itself. In that case, this can lead to collusion, where the reviewing model allows a harmful action to proceed due to a preference for the original model’s output.

In this paper, we investigate strategies to mitigate harmful self-preference. We build on the self-recognition hypothesis of \citet{panickssery2024llm}: self-preference is partially caused by the judges being able to recognize themselves. Following this hypothesis, we seek to mitigate self-preference by reducing the judge's ability to distinguish its own outputs from others: if the judge cannot tell which answer is its own, it's less likely to be biased. We do so by applying black-box perturbations to evaluation candidates to obfuscate their authorship.

We show that perturbations as simple as synonym replacement for a few words can indeed reduce harmful self-preference. This suggests that self-recognition and self-preference is indeed reliant on superficial, stylistic cues such as word choice. However, we also discover that a complete mitigation by black-box perturbations is challenging: when we apply further perturbations to neutralize the stylistic cues, self-preference recovers. 
The reason why this happens, as we design experiments to validate, is that self-recognition and self-preference happen on many semantic levels: when stylistic cues are removed, the judge can still distinguish their own answer from others when they express different opinions. This observation reveals the impossibility of a complete elimination of self-preference, which would require the judge's decisions to be decoupled from their beliefs about what is correct. A more plausible framing of the research question, as we discuss, should focus on preventing the judge from overly relying on its prior belief.
% The key challenge, as we discuss, is to formalize theories and collect empirical evidence for why we can expect the LM to behave differently---more truthfully and aligned with human values---when judging answers rather than giving them.

\section{Evaluating Harmful Self-preference}

We focus on pairwise comparison, a common format of using LM as a judge for benchmarking \cite{chiang2024chatbot} and reward modeling \cite{stiennon2020learning}.
Given answers from two LMs, the LM judge picks the better one according to criteria given in the prompt. When one of the LMs being evaluated is the same as the judge, we say that the judge is performing a self-evaluation.\footnote{For simplicity, we say that the judge is comparing its \textit{own} answer against a competitor. But we should note that the model receives different prompts in its two roles and does not behave exactly the same.}

We define self-preference as the judge selecting their own answer in self-evaluation. Such a preference is harmless if the judge's answer is indeed the better one, but harmful if otherwise. On tasks where answer quality can be objectively determined (e.g., by expert annotation), we can label self-preference as harmful when the judge selects their own answer when the competitor's answer is objectively better.

\subsection{Modeling}

We evaluate self-preference using five instruction-tuned LMs of varying sizes and capabilities: (1) Llama-3.1-8B, (2) Qwen2.5-7B, (3) Llama-4-Scout-17B, (4) Llama-4-Maverick-17B, and (5) DeepSeek-V3-37B. We set the temperature to zero for all judge decisions. 

\subsection{Testbed: Long-document QA}

To evaluate objective answer quality, we rely on multiple-choice questions with human-annotated correct answers. In particular, we use the QuALITY dataset \cite{pang2021quality}, which consists of long-form English passages (averaging around 4200 words) accompanied by multiple-choice questions. The passages are drawn from fictional narratives and magazine articles, designed to evaluate LM's comprehension of long-form texts. Each passage is followed by a question with four answer choices. We use the validation set of QuALITY, which contains 2,086 examples.

For each of the two models we are comparing, we provide the passage, the question, followed by the answer choices. We instruct the model to select an option from the list and provide a reasoning for the answer. We refer to these two parts of the model output as \textit{label choice} and \textit{reasoning}. We instruct the judge to evaluate the two models based on both their reasoning and their label choices, and provide context, including the passage and the question. Finally, evaluate the judge's decision based on whether the selected label choice is correct. Doing so allows us to objectively evaluate the judge and to label self-preference as harmful or harmless.

To control for ordering bias, we query the judge twice for each pairwise comparison, where the two evaluation candidates are swapped. A decision by the judge is only considered valid if it remains consistent across these two queries; otherwise, the judge is considered ambiguous. We report the frequency that a judge is ambiguous, and remove those decisions from subsequent analysis.

After removing ambiguous decisions, the judge's correctness---and subsequently the harmlessness of its self-preference---can still be ambiguous. Some answer pairs might contain two correct answers or two incorrect answers---or, more generally, contain two answers of objectively comparable quality. We also omit these examples from subsequent analysis.

To summarize, we use the following process to construct a dataset of answer pairs for our analysis: we first gather all models' answers on validation examples of QuALITY. Then, for every pair of models, we remove answer pairs with comparable answers. We then use every model as a judge to evaluate all remaining answer pairs, report the frequency of ambiguous decisions, and remove them. In this collection of answer pairs and corresponding judge decisions, we can always use the groundtruth label to objectively determine if the judge's decision is correct or not. We refer to this dataset as ``all examples''.

\subsection{Stronger models are more accurate judges}

\begin{figure}[ht]
\centering
\includegraphics[width=0.9\columnwidth]{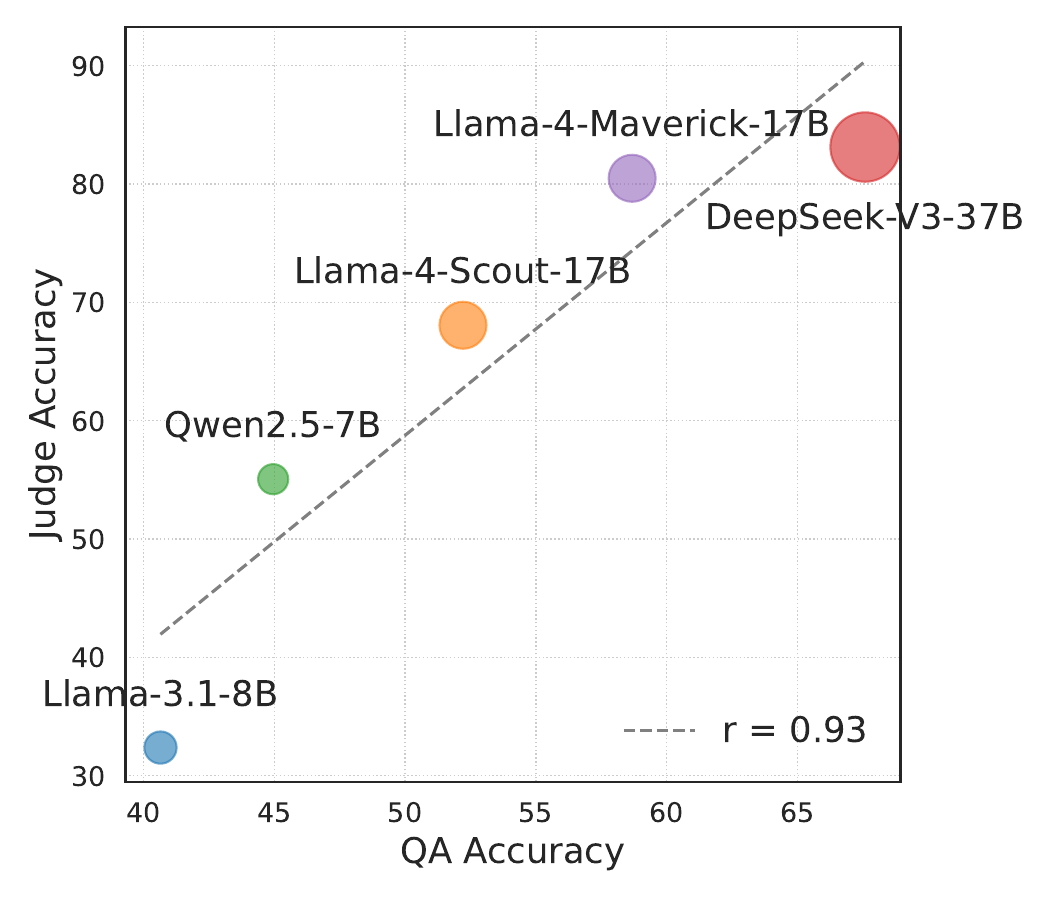}
\caption{Bigger models are more accurate at both answering questions and judging.}
\label{fig:qa_judge_acc}
\end{figure}

\begin{figure}[ht]
\centering
\includegraphics[width=0.9\columnwidth]{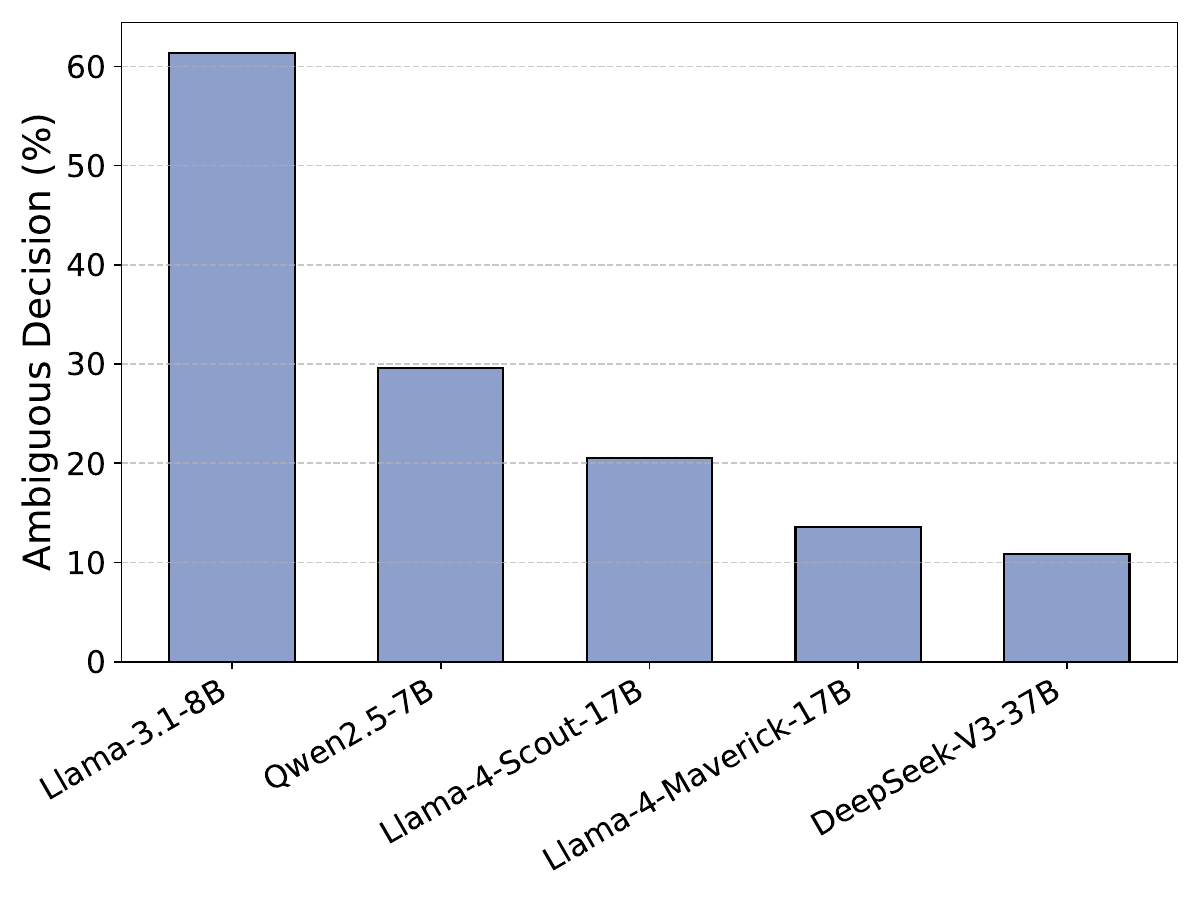}
\caption{Capable models are less sensitive to the order of evaluation and make fewer ambiguous decisions.}
\label{fig:decisive}
\end{figure}

Bigger models are generally more accurate at both answering questions and judging (Figure~\ref{fig:qa_judge_acc}) and make fewer ambiguous decisions (Figure~\ref{fig:decisive}).

In our experiments, we also find that more decisive models tend to be more accurate as judges (Figures \ref{fig:qa_judge_acc} and \ref{fig:decisive}), suggesting that when a model is confident about an answer selection, the decision is more likely to be correct.

In the following analysis, we omit all ambiguous answer pairs and judge decisions, such that for every pairwise comparison, we can definitively label the judge's decision as correct or incorrect. Because judges might not be ambiguous on the same set of examples, this omission leads to some variations in the answer pairs that each judge is evaluated on. 
% In Section~\ref{sec:variations} we discuss how this affects our findings.

\subsection{Stronger models overestimate their accuracy}

\begin{figure}[ht]
\centering
\includegraphics[width=0.9\columnwidth]{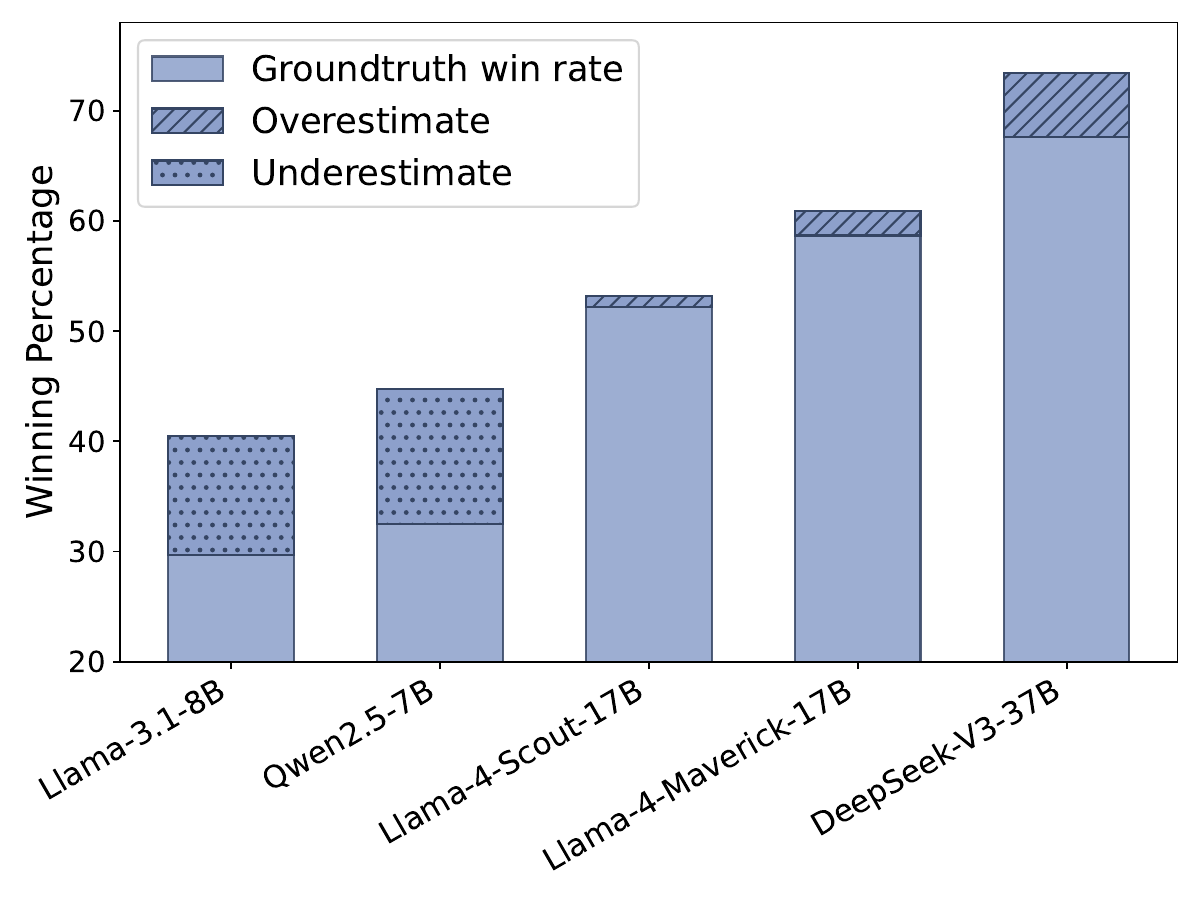}
\caption{Win rate of each model against all others as judged by the groundtruth compared to the model itself. Stronger models overestimate their accuracy; weaker models do the opposite.}
\label{fig:inflation}
\end{figure}

For every model, we calculate its win rate against all other models: the percentage of examples where the model provides an objectively better answer than another model.
We can compare the win rate calculated based on the groundtruth with the model's own estimate to characterize self-preference.
As we see in Figure~\ref{fig:inflation}, stronger models significantly overestimate their own accuracy. In particular, according to DeepSeek-V3's own estimation, its win rate against other models is 73\%, but in reality, its win rate is 66\%. Conversely, weaker models underestimate their accuracy: they frequently prefer other models' reasoning, even when their own answer is correct.

\begin{figure*}[t]
\centering
\includegraphics[width=0.9\textwidth]{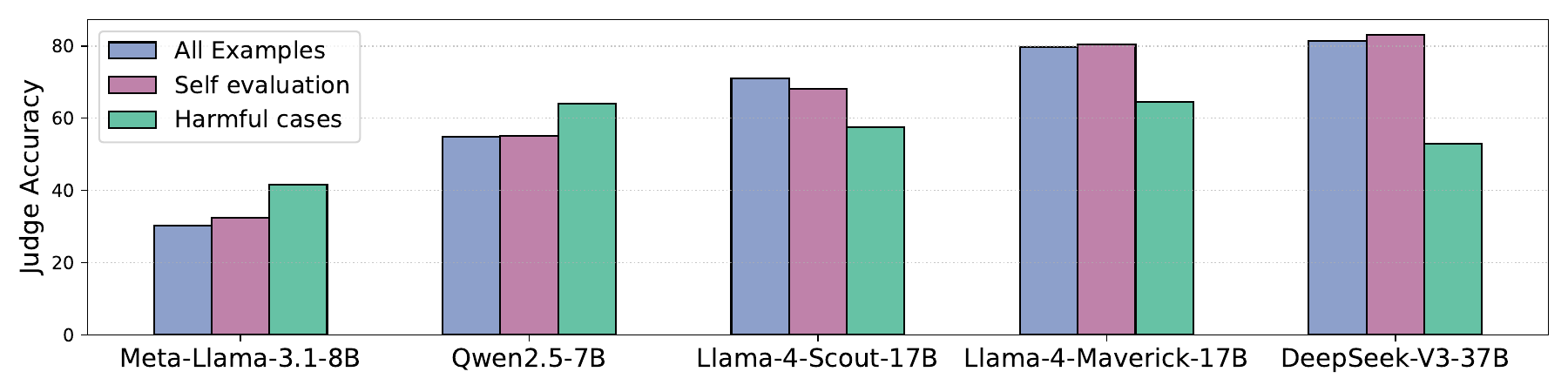}
\caption{Strong models are significantly less accurate on examples where their own answers are wrong (harmful cases), but have a higher overall judge accuracy.}
\label{fig:accuracy_self_eval_loss}
\end{figure*}

\subsection{Harmful self-preference in large models}

To further isolate the impact of harmful self-preference, we zoom in on each judge's accuracy in self-evaluation and in the harmful cases---the subset of self-evaluation where the judge's answer is wrong. As shown in Figure \ref{fig:accuracy_self_eval_loss}, the accuracy of more capable judges takes a bigger dive in the harmful cases, suggesting more significant harmful self-preference. In particular, DeepSeek-V3 is more accurate than LLama-4-Scout-17B overall and in self-evaluation, but its degradation on harmful cases is much more significant, making it less accurate than the two LLama-4 models in those cases. This indicates that DeepSeek-V3, as a judge, shows a stronger tendency to stick to its own reasoning when it's wrong---it is more ``egotistical''. Harmful self-preference is distinctively a tendency that occurs in the three larger models: although they are overall more accurate judges, we can trust them less to spot their own mistakes, even when a better alternative is presented.

\section{Mitigating Harmful Self-preference}

The previous section validates that despite higher accuracy, the more capable judge also exhibits harmful self-preference. This is concerning: in general, we will always want to use the most capable and accurate judge, but the harmful self-preference also means that we need to be more cautious with them. In this section, we investigate strategies to mitigate the self-preference bias, and empirically examine whether there is a trade-off between accuracy and bias. We base our study on the self-recognition hypothesis that self-preference is partly driven by the judge's ability to differentiate their own answers from others. Motivated by how LM detection methods~\cite{mitchell2023detectgpt} exploit the model's sensitivity to small changes in the input, we hypothesize that self-recognition is also sensitive to small perturbations, which means we can mitigate self-preference by reducing self-recognition using small perturbations.

\subsection{Validating the self-recognition hypothesis}

% \begin{figure}[h]
% \centering
% \includegraphics[width=0.9\columnwidth]{figures/self_recog_corr_harmful.pdf}
% \caption{Pearson correlation between self-recognition and self-preference across models. Stronger models generally exhibit a higher correlation, suggesting their self-awareness in recognizing and preferring their own responses is more consistent compared to smaller or weaker models.}
% \label{fig:correlation_recog_pref}
% \end{figure}

We begin our investigation by validating the connection between self-preference and self-recognition. Following \citet{panickssery2024llm}, we prompt the judge to identify which of the two evaluation candidates it believes to have been generated by itself, in a context separate from self-evaluation. 
In the subset of harmful cases, there was a positive correlation between self-recognition and self-preference, $r(6667) = 0.63$, $p < .001$, 95\% CI [0.62, 0.65].
A chi-square test further finds a statistically significant relationship between them ($p$ \textless 0.001, df = 1). This indicates that models are substantially more likely to favor incorrect responses that they recognize as their own.

% \begin{table}[h]
% \centering
% \small
% \begin{tabular}{lcc}
%  & Self-Preference & No Self-Preference \\
% Self-Recognition   & 471  & 218  \\
% No Self-Recognition      & 239  & 5563 
% \end{tabular}
% \caption{ in the harmful case based on table showing the relationship between self-recognition and preference.}
% \label{table:contingency}
% \end{table}

\subsection{Perturbation by synonym replacement}
\begin{figure*}[t]
\centering
\includegraphics[width=0.8\textwidth]{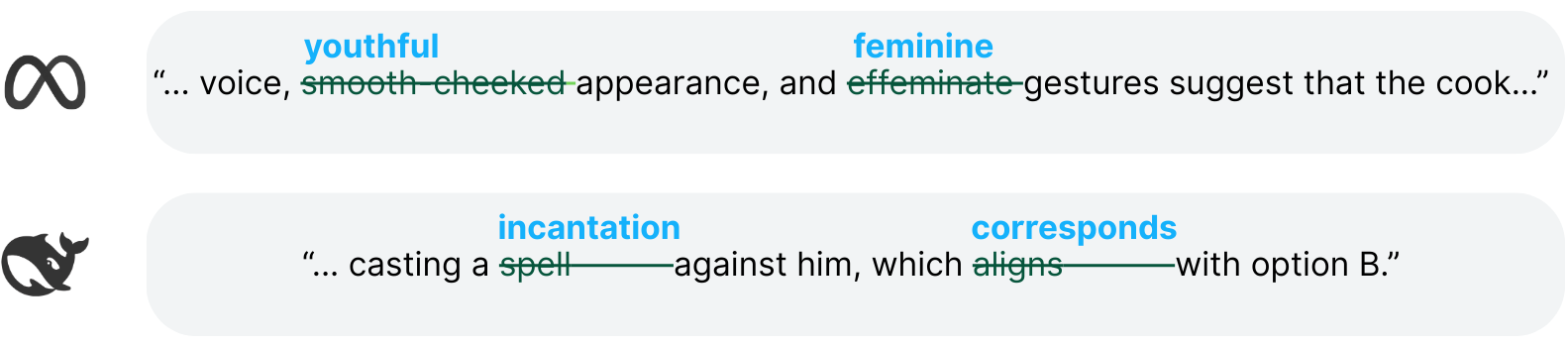}
\caption{Examples of synonym replacement (top, Llama-3.1 and bottom DeepSeek-V3). Objective answer quality is not affected.}
\label{fig:differences_perturb_samples_words}
\end{figure*}
Based on our validation that self-preference is associated with self-recognition, we explore strategies to mitigate self-preference by obfuscating authorship: if the judge cannot recognize the answer as its own, it wouldn't favor it disproportionately.

One key intuition drawn from the LM detection literature is that LMs are more sensitive to perturbations when they are applied to texts that are generated by the same LM than others. Zero-shot methods \cite{mitchell2023detectgpt,bao2023fast} achieve high detection accuracy by drawing a threshold on the perplexity change when small perturbations are applied to the text being evaluated: if the perturbation leads to a large change in perplexity, it's likely that the text was generated by the LM.
Following this intuition, we hypothesize that self-recognition is also highly dependent on stylistic cues such as word choices, and that small perturbations with no affect on the semantic meaning are sufficient for reducing self-recognition.

The first perturbation we experiment with is synonym replacement: in self-evaluation, we replace two words in \textit{the judge's answer} with their synonyms. We anticipate that this would reduce the judge's ability to discern which of the two evaluation candidate is its own generation, and subsequently reduce self-preference.

We prompt a LLaMA-3.3-70B (not in our judge pool) to identify replacement candidates in the reasoning that are neither stop words nor present in the question, to avoid disruptive changes and ensure the relevance of the answer remain unaffected. We use a third-party model for this purpose to avoid bias against any judge.
This process produces subtle word changes that do not affect answer quality or relevance to the question (Figure~\ref{fig:differences_perturb_samples_words}).
We manually verify that by looking at the perturbed answer, we cannot tell which word has been replaced; even when comparing the original and perturbed versions side by side, it is not possible to determine which version is which.
The number of words to replace is a hyperparameter that we leave for future work to explore.

\subsection{Synonym replacement reduces self-recognition and self-preference}

\begin{figure}[t!]
\centering
\includegraphics[width=0.9\columnwidth]{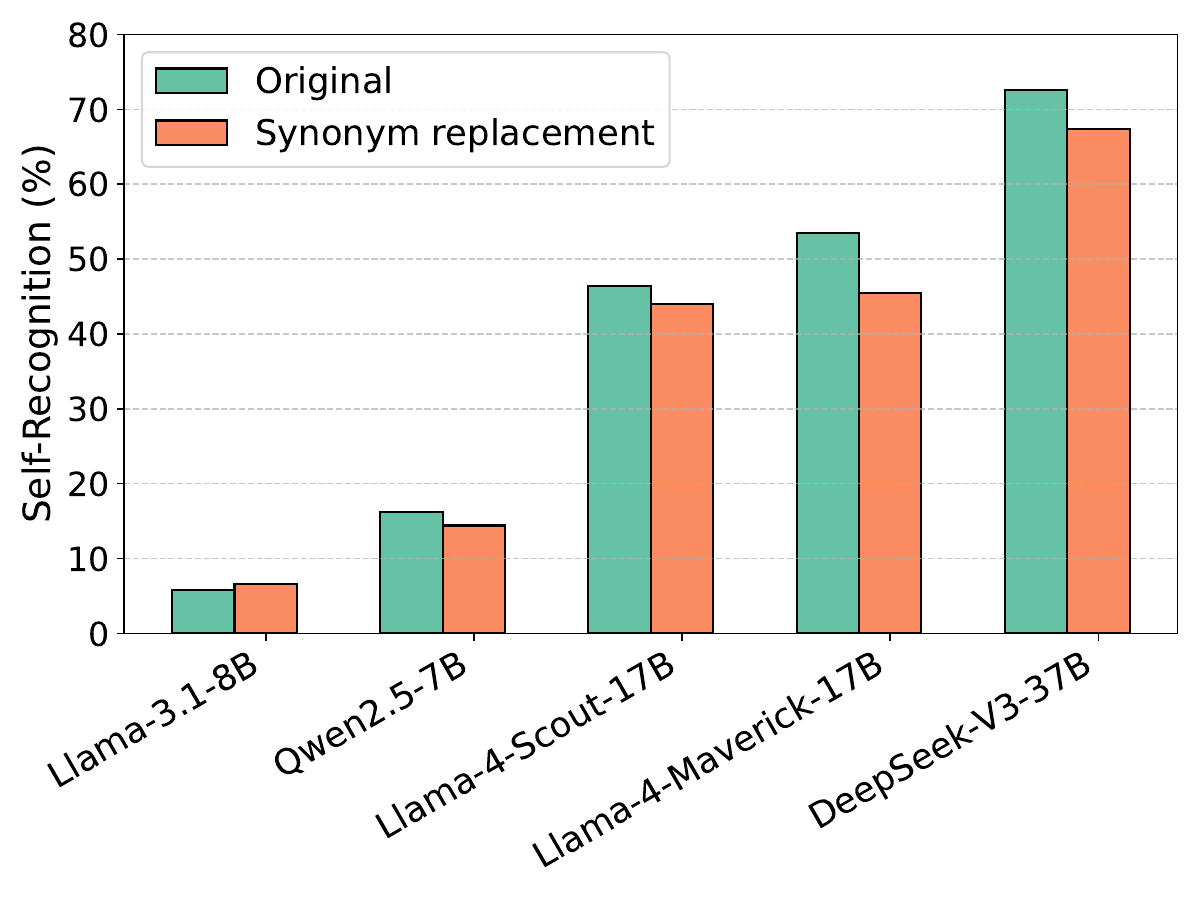}
\caption{Two-word synonym replacement reduces overall self-recognition, especially for the more capable models.}
\label{fig:self_recognition_perturb}
\end{figure}

\begin{figure}[t!]
\centering
\includegraphics[width=0.9\columnwidth]{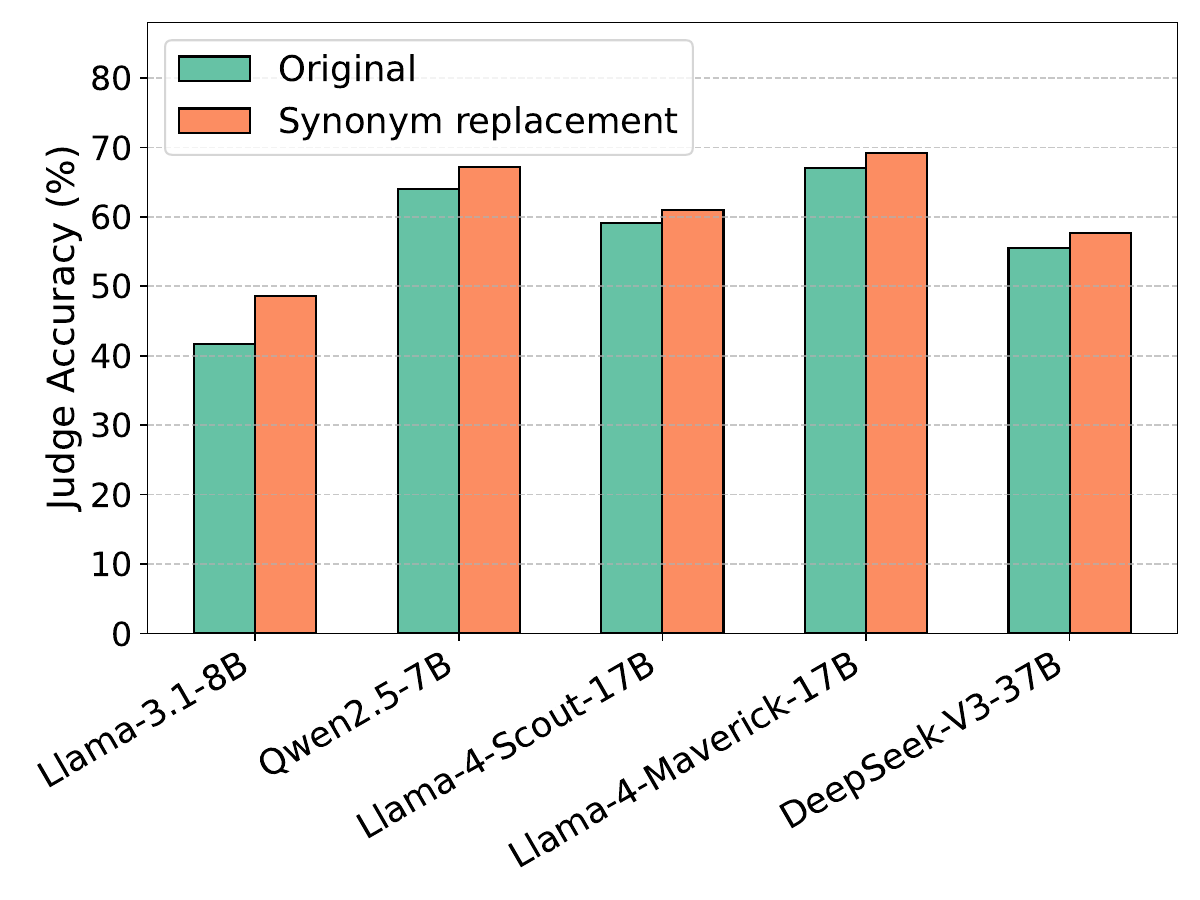}
\caption{Synonym replacement reduces harmful self-preference and increases the accuracy of judges when their own answer is wrong. A paired t-test revealed a significant difference in preference between the original and perturbed conditions ($p$ \textless.01).}
\label{fig:lose_case_perturb_accuracy_token}
\end{figure}

We repeat the self-recognition evaluations on the harmful cases and observe a significant reduction in self-recognition after perturbing only two words (Figure \ref{fig:self_recognition_perturb}), with a more pronounced impact on the stronger models. This confirms our hypothesis that self-recognition is sensitive to superficial, stylistic cues like word choices, and that black-box perturbations can effectively obfuscate authorship.

We then validate that perturbations achieve the expected effect of mitigating self-preference. Figure~\ref{fig:lose_case_perturb_accuracy_token} shows a consistent increase in the judge's accuracy on the harmful cases---examples where the judge's answer is objectively the worse one.
This means that after perturbation, the judges are indeed less likely to prefer their own incorrect answers---a reduction in harmful self-preference.

% We also validate that similar trends exist across 4 other models from a diverse family when looking at the data from our original set of 5 models (see Appendix for details).

\subsection{Judges are more affected by perturbations in self-evaluation than as a third-party}

% \begin{figure}[h]
% \centering
% \includegraphics[width=0.9\columnwidth]{figures/pref_comparison_self_thirdparty_aggregate_token.png}
% \caption{An aggregated comparison of the impact of self-evaluation versus third-party evaluation on the same subset of lose-case data. Across both perturbation types, self-evaluation exhibits a larger shift in preference. This indicates that the objective quality of responses remains largely unaffected by perturbations, with changes primarily influencing self-evaluating models.}
% \label{fig:pref_change_aggregate}
% \end{figure}

\begin{figure}[t]
\centering
\includegraphics[width=0.9\columnwidth]{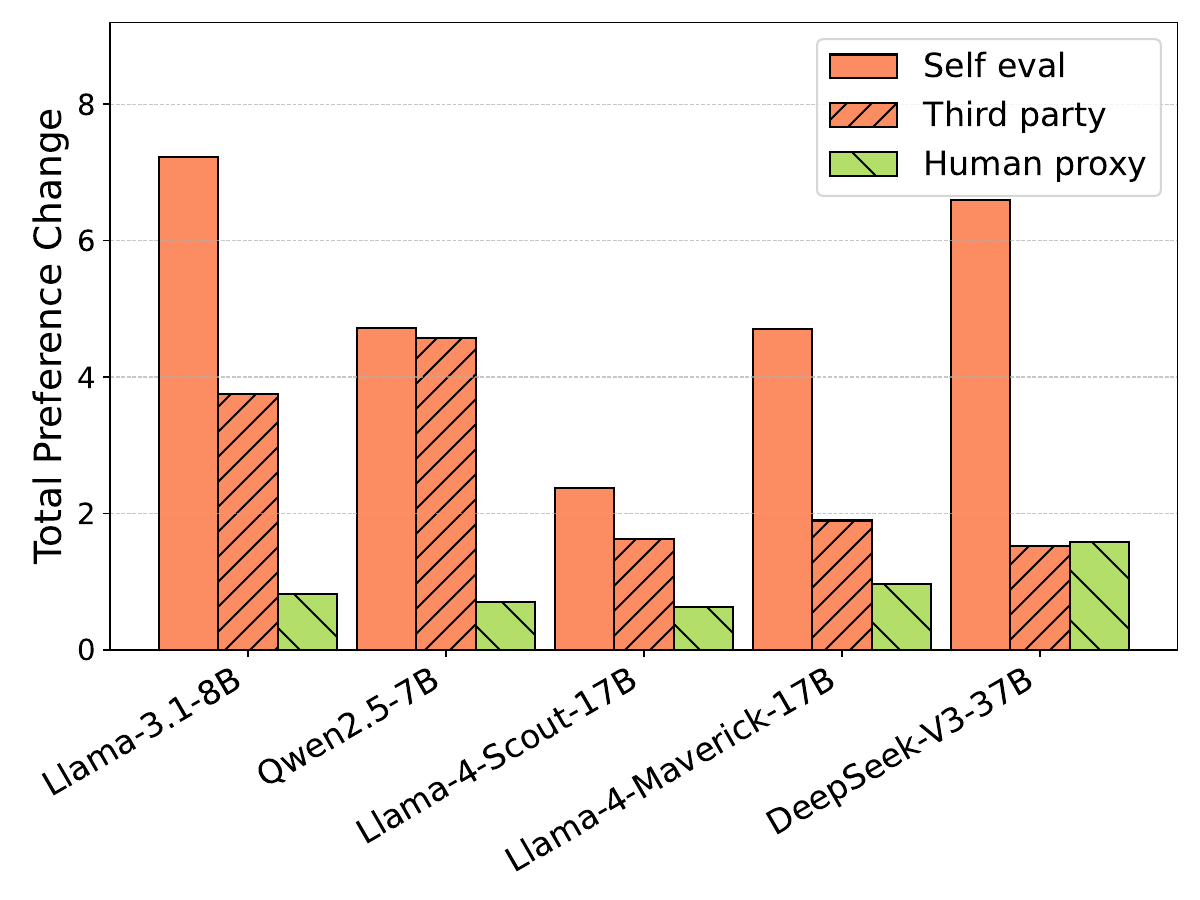}
\caption{The impact of perturbation measured by total preference change is significantly higher on judges in self-evaluation than as a third-party. Human proxy also confirms that perturbation has a minimal effect on answer quality.}
\label{fig:judge_centric_view}
\end{figure}

Despite our effort to minimize the impact of synonym replacements on the answer's semantic qualities, we cannot rule out that it contributes to the reduction of self-preference, that the perturbed answer is, in fact, lower quality.
In other words, the perturbations can lower self-preference through two paths: by reducing self-recognition as we desire, and by damaging the answer's quality.
To control for the latter, we examine the impact of perturbation on each judge as a self-evaluating judge and as a third-party judge.
To be more concrete, we first compute the total change in preference of all judges in self-evaluation, as we do for Figure~\ref{fig:lose_case_perturb_accuracy_token}; we then use each judge to evaluate answer pairs generated by two other models, and compute the total preference before and after perturbation.
The underlying answer pairs (including the perturbed versions) remain the same under these two conditions; the only variable is whether the judge's own answer is being perturbed.
As we see in Figure~\ref{fig:judge_centric_view}, all judges are significantly more sensitive to perturbations in self-evaluation than as a third-party---although synonym replacement might have a small impact on objective answer quality, the reduction in self-preference is mainly caused by reducing self-recognition.
We additionally validate that perturbations have minimal effect on objective quality using a frontier commercial model (o3) as an approximation for human judgment. To complement this, we also conduct a human study on 500 examples, confirming that human raters judged the original and perturbed versions as equivalent in quality.

% \begin{figure}[h]
% \centering
% \includegraphics[width=0.9\columnwidth]{figures/third_party_o3.pdf}
% \caption{o3, as a human-proxy shows minimal changes after synonym replacement on the loss-case data.}
% \label{fig:third_party_o3}
% \end{figure}

% Figure \ref{fig:self_eval_perturb_changes} illustrates the effect of applying synonym replacement to response A (the judge's own answer) versus response B (the competitor's answer). Although perturbing A generally leads to a larger change in preference (as shown in Figure \ref{fig:pref_change_aggregate}, we observe that two models—Qwen2.5 and Llama-4-Scout—are more sensitive to perturbations in the competitor's response, indicating higher variability in their judgment behavior.

% \begin{figure}[t]
% \centering
% \includegraphics[width=0.9\columnwidth]{figures/self_eval_perturbA_perturbB.pdf}
% \caption{Self-evaluation is more sensitive to perturbations applied to the model’s own response, with Qwen2.5 and Llama-4-Scout exhibiting greater variability and less consistent behavior under such conditions.}
% \label{fig:self_eval_perturb_changes}
% \end{figure}

\subsection{Perturbation by Judge-Paraphrasing}

\begin{figure*}[t]
\centering
\includegraphics[width=0.9\textwidth]{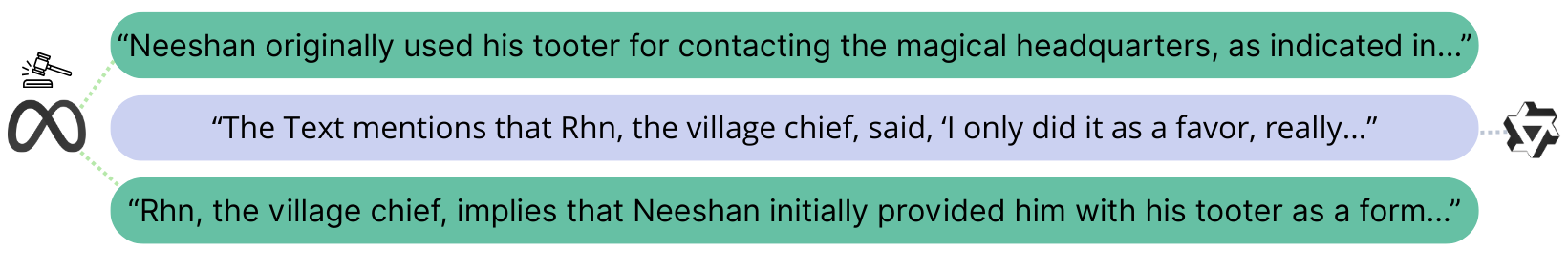}
\caption{Examples of judge paraphrasing. We prompt the judge (in example, Llama-4-Scout) to paraphrase the competitor's (Qwen-2.5) answer while maintaining semantics, so that both evaluation candidates look like they were produced by the judge in terms of style.}
\label{fig:differences_paraphrase_samples}
\end{figure*}

Our experiments with synonym replacement validate that stylistic cues such as word choices are indeed the basis of self-recognition, and that by perturbing those, we can reduce self-preference. So we take it further: we use the judge to paraphrase the competitor's answer so they both look like they were produced by the judge. If our logic holds, this should completely neutralize these superficial cues and further reduce self-preference. We call this perturbation \textit{judge paraphrasing}, and Figure \ref{fig:differences_paraphrase_samples} presents a few examples.

\begin{figure}[ht]
\centering
\includegraphics[width=0.9\columnwidth]{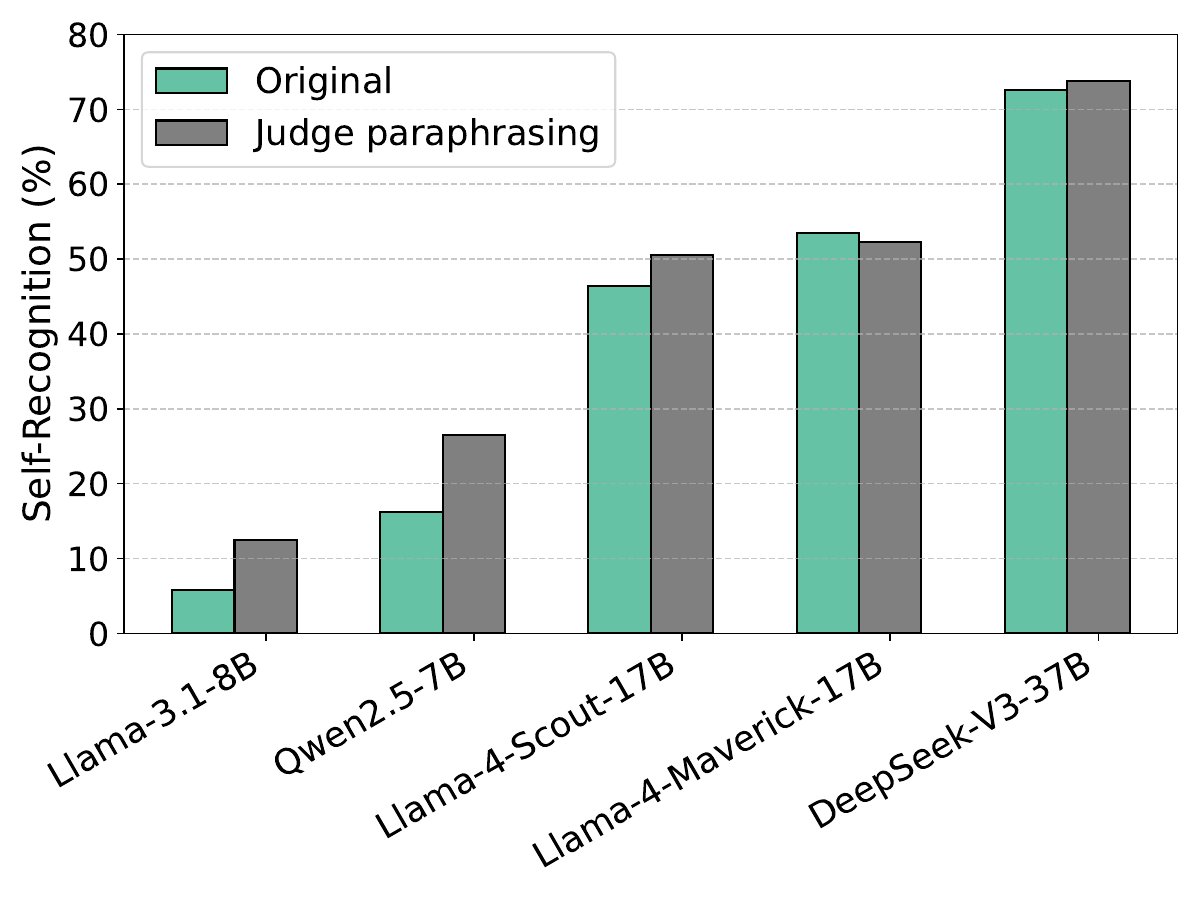}
\caption{Using the judge to paraphrase the competitor---which should neutralize all stylistic cues---ends up increasing self-recognition.}
\label{fig:self_recognition_paraphrase}
\end{figure}

\subsection{Judge paraphrasing increases self-recognition and self-preference}

\begin{figure}[ht]
\centering
\includegraphics[width=0.9\columnwidth]{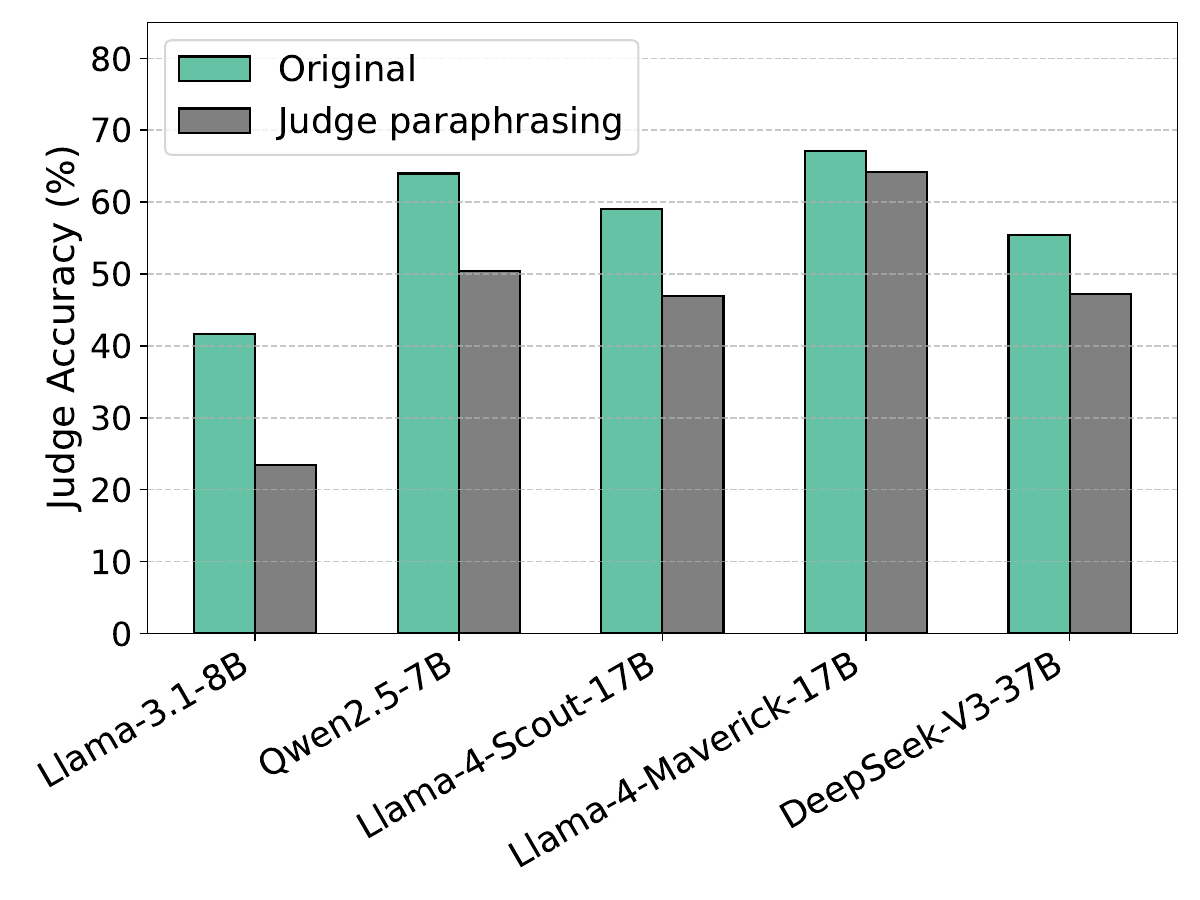}
\caption{Judge paraphrasing decreases judge accuracy in the harmful cases and exacerbates harmful self-preference.}
\label{fig:lose_case_perturb_accuracy_parphrase}
\end{figure}

\begin{figure}[t!]
\centering
\includegraphics[width=0.9\columnwidth]{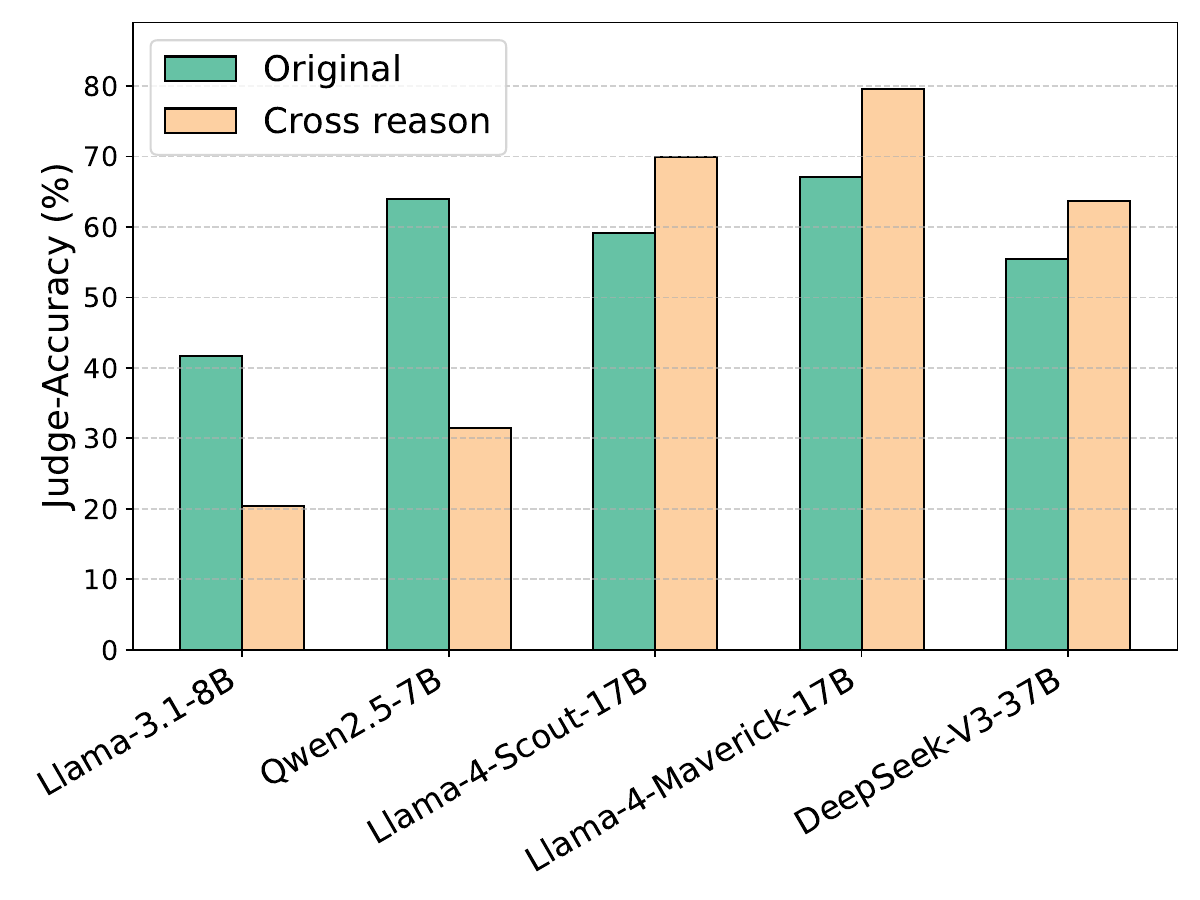}
\caption{Using the judge to generate a reason for the competitor's option and vice versa yields 2 observations: weaker models tend to prefer the competitor's reasoning more, and stronger models tend to choose their own reasoning.}
\label{fig:lose_case_shared_belief}
\end{figure}

Judge paraphrasing achieves the opposite effect than we expected: it increases self-recognition (Figure~\ref{fig:self_recognition_paraphrase}), and increases harmful self-preference (Figure~\ref{fig:lose_case_perturb_accuracy_parphrase}).
This means that, when the stylistic differences are removed, the judge is relying on semantic differences between the two answers to judge which one was written by itself and which one to favor. In retrospect, this makes sense: now that style is no longer a factor, it is simply choosing the answer that it ``agrees'' with.

Contrasting the two perturbations, we see that self-recognition and self-preference can occur at multiple semantic levels: it can prefer an answer because its style resembles the judge's own, or because it expresses an opinion that the judge agrees with. Based on our experiments, when \textit{superficial self-recognition} is removed, \textit{shared belief} takes over as the driving force behind self-preference.

To examine the dynamics between these two factors, we cross-examine each judge with answer pairs where these two factors act against each other: the label/option chosen by the judge paired with a reasoning generated by the competitor, and vice versa. In other words, the judge would see one answer expressing an opinion that it agrees with, and another answer with an opposing opinion, but in a style it's more familiar with. If, after this intervention, the judge changes its decision to favor the answer with a familiar-looking style, we can conclude that superficial self-recognition has a higher weight in that decision.

Figure \ref{fig:lose_case_shared_belief} illustrates the two contrasting outcomes observed in the cross-reason experiments. For the weaker models, we observe a negative effect of shared belief, where incorporating the competitor’s reason reinforces the model’s original, incorrect answer, ultimately decreasing judge accuracy. In contrast, all three stronger models exhibit a positive self-recognition effect: they are more inclined to select the competitor’s (correct) answer when it is paired with an explanation generated by themselves. 
This observation reinforces our finding that stronger models have a higher self-recognition ability and prefer answers written by them, even if it goes against their original selection.
% However, the magnitude of preference change is more pronounced in weaker models when compared to the larger models, which reinforces the finding that they are more reliable judges.

% \section{Does an Ensemble of Judges provide better objective evaluation?}

% We explore the effectiveness of a majority voting ensemble \cite{zhao2024language} in which all five models provide pairwise preference judgments for a given pair of answers. In this framework, two models participate in self-evaluation, while the remaining three act as third-party judges. Only decisive judgments are included in the voting process, and ambiguous results are discarded. For each target model, we evaluate performance on a consistent subset of loss case samples and compare ensemble-based outcomes with those from self-evaluation. As illustrated in Figure \ref{fig:ensemble}, ensemble voting leads to improved overall evaluator accuracy, with the most substantial improvements observed in loss case samples for weaker models.

% \begin{figure}[t]
% \centering
% \includegraphics[width=0.9\columnwidth]{figures/ensemble.pdf}
% \caption{Ensemble of all judges in the loss case samples. The average accuracy increases from 56.1\% to 80.6\%.}
% \label{fig:ensemble}
% \end{figure}

\subsection{Mitigating Self-Preference in Coding Task}

\begin{figure}[ht]
\centering
\includegraphics[width=0.99\columnwidth]{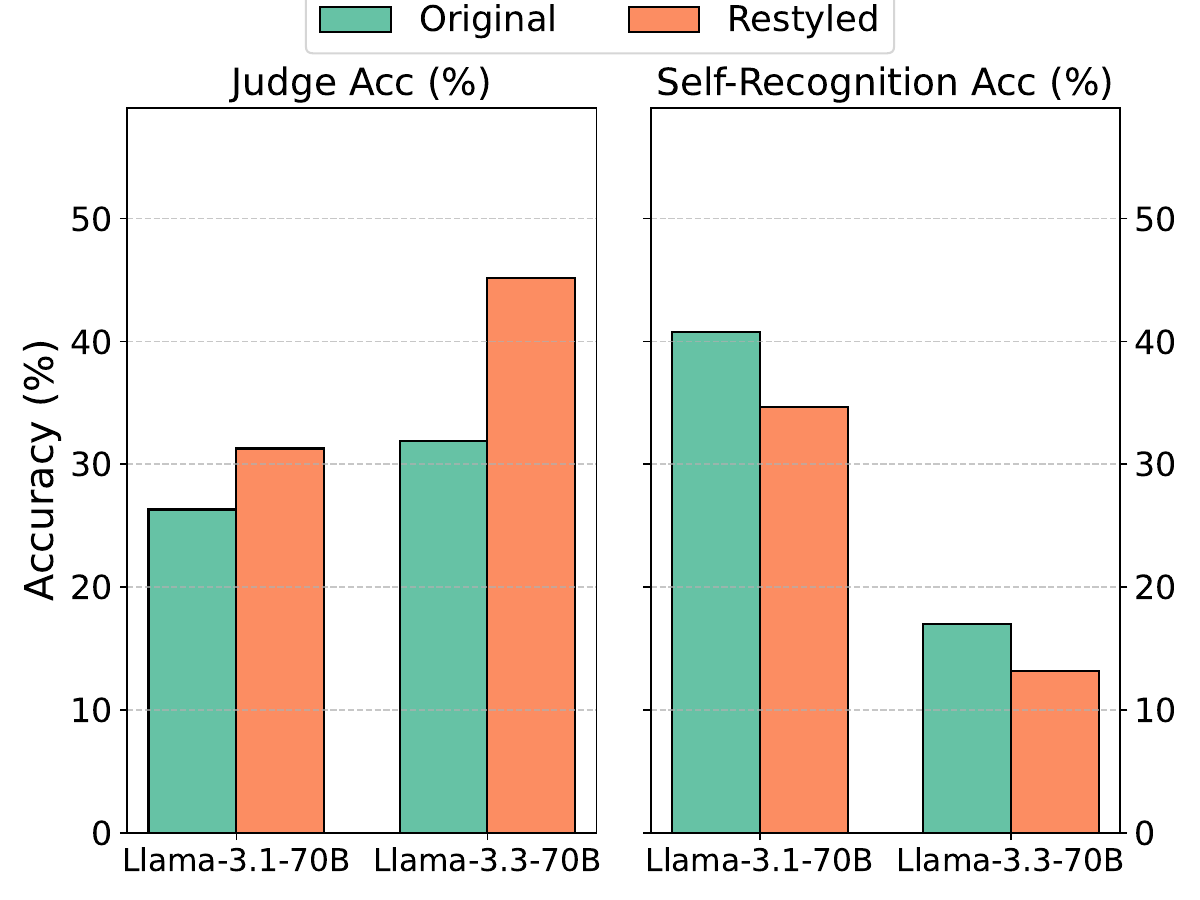}
\caption{Restyling the code improves the judge-accuracy in harmful cases, while reducing self-recognition ability.}
\label{fig:self_recognition_judge_acc_coding}
\end{figure}

To examine whether our mitigation of self-recognition and self-preference extends beyond a single task, we use the MBPP+ dataset from \cite{chen2025llm} to test code-level obfuscation techniques. Specifically, we use coding solutions generated by Llama-3.1-70B and Llama-3.3-70B, then evaluated against outputs from six other models. We rewrite the original code to introduce idiomatic style variations (e.g., converting imperative structures into declarative ones) while ensuring functional equivalence by matching the original code for all test cases provided in the question. Figure \ref{fig:self_recognition_judge_acc_coding} illustrates the overall trend observed across both judge models: introducing minor stylistic variations while preserving the underlying code logic consistently reduces self-recognition ability and improves judge accuracy.

%-----------------------------

\section{Conclusion, Discussion, and Limitations}

We study self-preference in LM judges using QuALITY, a long-document question answering dataset. We find that the more capable models exhibit significant harmful self-preference: although they are generally more accurate, we need extra caution when tasking them to spot their own mistakes, even when objectively better options are presented. We validate the contribution of self-recognition to such bias, and present viable mitigation strategies using inference-time perturbations. At the same time, we illustrate the complexity of this task: self-recognition and self-preference can occur at multiple semantic levels, and a complete elimination is not plausible---developing mitigation strategies requires a more nuanced approach to formulate the objective. 

\paragraph{White-box mitigation} Recent work explores white-box methods such as activation steering to manipulate self-recognition \cite{ackerman2024inspection} and self-preference \cite{nguyen2025manipulating}. We focus on black-box methods for two main reasons. First, the processes are less dependent on having access to specific models' internals and are more applicable as a standardization process for benchmarks. Second, the perturbations allow us to confirm the hypothesis that self-recognition and subsequently self-preference are indeed sensitive to superficial cues. An important future work would be to compare white-box and black-box methods for their robustness and flexibility, both of which are crucial to the proper debiasing of LM judges.

\textbf{Models choose what they think is theirs}  Our findings suggest a strong connection between self-recognition and harmful self-preference—models tend to favor the output they believe they generated. Weaker models generally struggle with self-recognition and, as a result, exhibit weaker self-preference. This limitation can be advantageous: for instance, our experiments show that Qwen2.5 performs on par with models nearly five times larger in lose-case evaluations precisely because it does not demonstrate harmful self-preference.

\textbf{Recommendations for LLM-as-a-Judge Systems} We suggest several strategies to enhance the reliability of LLM-based judge systems. First, since models tend to favor their own outputs, excluding the evaluated model from serving as a judge can help reduce this bias. Additionally, targeted token perturbations can eliminate superficial self-identifying cues without compromising the overall quality of the response. We encourage future works to explore other methods of authorship obfuscation. Among the most effective approaches, employing a majority-vote ensemble of decisive judges offers a robust means of mitigating preference bias and improving judgment fairness.

\textbf{Harmful Self-Preference and scalable oversight} LLM-based evaluations is a key to achieving scalable oversight in many complex domains, where human supervision is costly and limited (specialized). A stronger model may be used as a supervisor in many tasks. Our experiments show that stronger models are able to recognize themselves more frequently, and suffer the most in loss cases due to harmful self-preference. Thus, supervisor models in scalable oversight problems must be explored in these tasks, where the models are not as effective as they are likely to provide biased or incorrect supervision. One such example is using an LLM as a monitor for reward hacking or scheming output, where using the same model for the detection of  ``cheating" may result in undetected examples.

\section{Limitations}

\textbf{Controlling for Confounding Biases}. While we focus on an objective task setup—where each evaluator is presented with a single correct answer and is instructed to judge based on the accompanying rationale—we do not explicitly control for factors such as persuasiveness, stylistic tone, or length bias that may influence the model’s decision. These attributes can unintentionally sway preferences, even when the reasoning is logically sound. As a result, our analysis captures a potential interplay between subjective and objective factors. Future work should explore controlled manipulations of these stylistic elements to more rigorously assess the fairness and robustness of LMs in judgment roles.

\textbf{Broader Impacts of Perturbation}. Our study primarily targets the mitigation of self-preference bias in loss-case scenarios. While the perturbation techniques we evaluate are effective in this context, their impact on win-cases and on settings where both reasons are perturbed remains unexplored. A more comprehensive evaluation across diverse scenarios is necessary to determine whether these methods can serve as a general strategy for addressing superficial self-recognition and associated biases in model judgment.

% *Diversity Across Model Families*. #text(red)[The results are removed] #text(blue)[Although our findings provide preliminary evidence of harmful self-preference within models from the same family, our analysis is limited to three Llama-based models.] To more thoroughly assess whether such preferences arise from shared architectural features or overlapping training data, a broader inclusion of models from diverse families is necessary. Future work could adopt a family-based ensemble framework—where evaluators belong to the same model family—to further investigate whether biases toward outputs from related models persist, even in objectively grounded tasks.

%
%
%
%
%
%
%
%

\bibliography{aaai2026}

\clearpage

\appendix
\section*{Appendix}
\label{appendix}

The appendix is organized into six sections: (1) implementation details of the experiments, (2) results with additional judge models on the QuALITY dataset, (3) ablation studies of alternative obfuscation methods, (4) human-control group study, (5) obfuscation and mitigation analysis in the coding task, and (6) all prompts used in the experiments.

\subsection{Implementation Details}
% Your content here
\paragraph{Models} We utilize the following model versions from together.ai's  serverless inference end-points: 
\begin{itemize}
    \item Llama models:meta-llama/Llama-3.1-8B-Instruct, meta-llama/Llama-4-Scout-17B-16E, meta-llama/Llama-4-Maverick-17B-128E-Instruct-FP8, meta-llama/Llama-3.3-70B-Instruct-Turbo, meta-llama/Llama-3.3-70B-Instruct
    \item Qwen models: Qwen/Qwen2.5-7B-Instruct, Qwen/Qwen3-Coder-480B-A35B-Instruct-FP8
    \item DeepSeek models: deepseek-ai/DeepSeek-V3
    \item Mistral models: mistralai/Mistral-Small-24B-Instruct-2501
    \item Moonshotai models: moonshotai/Kimi-K2-Instruct
    \item OpenAI models: openai/gpt-oss-20b
    \item Zai models: zai-org/GLM-4.5-Air-FP8
\end{itemize}

\paragraph{Response and Judgement generation} We use an asynchronous client with a semaphore limit of 10 to enable parallel generation of LLM outputs and verdicts. For answer generation, the model is prompted to return a response in JSON format, from which we extract the selected option (A–D) and the accompanying reason/justification. For pairwise preference evaluations, the model is instructed to return either `A' or `B' to indicate its preferred response. All LLM-generated answers and judgment outputs are available on our GitHub repository. 
% \footnote{Code available at https://github.com/Taslim-M/mitigatingselfpreference}

%----------------------------------------------
% \subsection{Data Distribution for QuALITY Tasks}

% \begin{table}[ht]
% \centering
% \caption{Comparison of Non-Ambiguous and Ambiguous Decisions Across Evaluators for all data. 
% Non-ambiguous decisions indicate consistent forward and backward judgments, 
% while ambiguous decisions denote conflicting evaluation outcomes.}
% \begin{tabular}{lcc}
% \toprule
% \textbf{Evaluator} & \textbf{Non-Ambiguous} & \textbf{Ambiguous} \\
% \midrule
% Llama-3.1-8B       & 1192 & 1893 \\
% Qwen2.5-7B             & 2022 & 851  \\
% Llama-4-Scout-17B         & 2242 & 578  \\
% Llama-4-Maverick-17B & 2001 & 314  \\
% DeepSeek-V3              & 2602 & 317  \\
% \bottomrule
% \end{tabular}
% \label{tab:ambiguity_results}
% \end{table}

%----------------------------------------------
\subsection{Verifying Additional Judge Models (QuALITY)}

Table \ref{tab:quality_additional_judge_accuracy} shows the impact of 2-word synonym replacement on judge-accuracy for additional judge models when evaluating self-preference against the 5 models in our initial pool. We explore a range of capable models from different families. We observe a reduction in harmful self-preference in the loss-case with improvement of judge-accuracy, as well as a reduction in overall self-recognition ability (Table \ref{tab:quality_addtiional_self_recognition}).

\begin{table}[ht]
\centering
\begin{tabular}{p{3.0cm} p{1.4cm} p{2.0cm}}
\toprule
\textbf{Judge} & \textbf{Original} & \textbf{Synonym-Replacement} \\
\midrule
Mistral-Small-24B & 26.78\% & 33.52\% \\
Kimi-K2 & 28.84\% & 32.18\% \\
GPT-OSS 20B & 28.28\% & 31.43\% \\
GLM-4.5-Air & 60.56\% & 63.94\% \\
\bottomrule
\end{tabular}
\caption{Judge Accuracy Results in Loss-Case}
\label{tab:quality_additional_judge_accuracy}
\end{table}

\begin{table}[ht]
\centering
\begin{tabular}{p{3.0cm} p{1.4cm} p{2.0cm}}
\toprule
\textbf{Judge} & \textbf{Original} & \textbf{Synonym-Replacement} \\
\midrule
Mistral-Small-24B & 8.28\% & 7.11\% \\
Kimi-K2 & 66.99\% & 65.89\% \\
GPT-OSS 20B & 76.23\% & 72.94\% \\
GLM-4.5-Air & 46.17\% & 42.84\% \\
\bottomrule
\end{tabular}
\caption{Overall Self-Recognition Accuracy}
\label{tab:quality_addtiional_self_recognition}
\end{table}

\subsection{Comparing Against Other Obfuscation Methods}

We leave the exploration of additional authorship obfuscation techniques to future work.
In this study, we present an ablation analysis of how four-word perturbation, spelling errors, and Tarot \cite{loiseau2025tarot} affect preference shifts in both self-evaluation and third-party evaluation.
We argue that obfuscation methods producing smaller changes in third-party evaluation, relative to self-evaluation, are more effective for mitigating self-preference.

\begin{figure}[h]
\centering
\includegraphics[width=0.95\columnwidth]{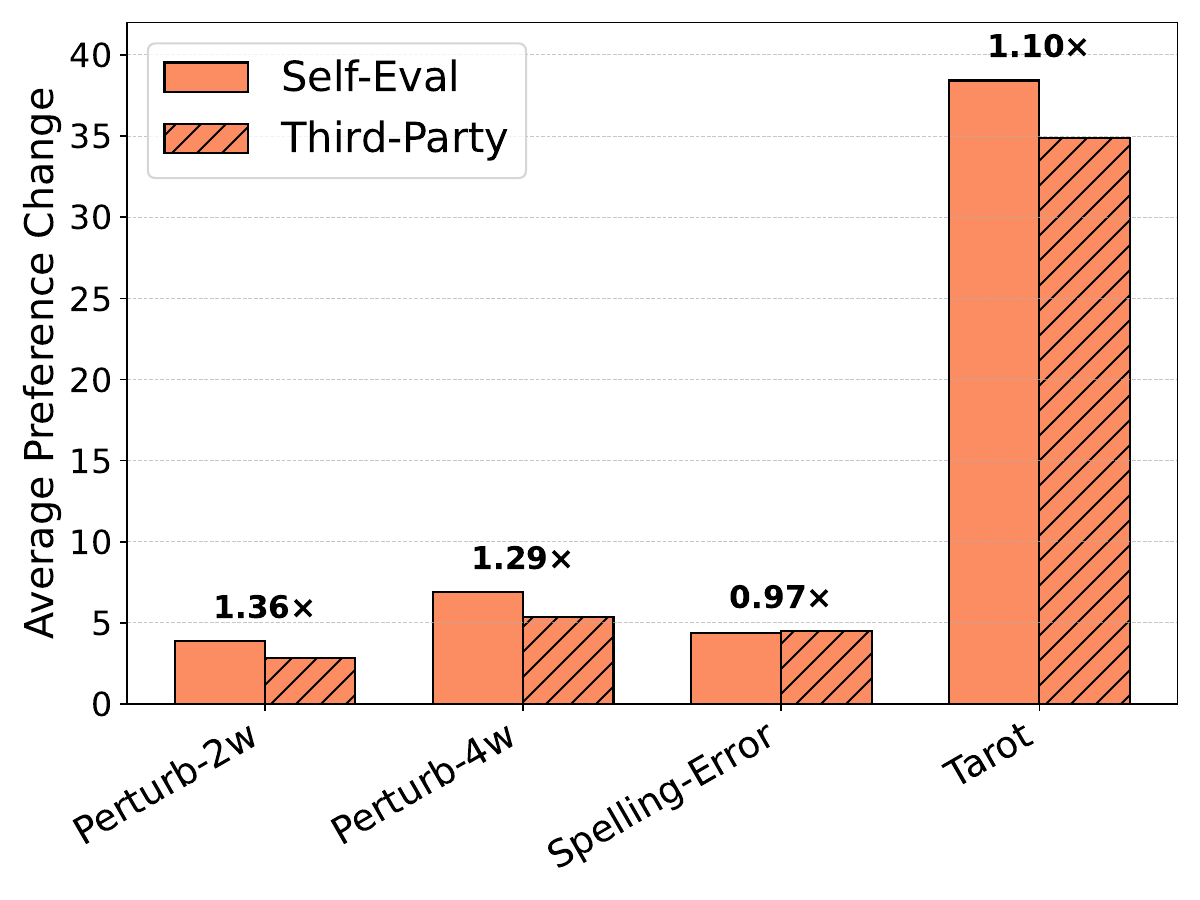}
\caption{Comparison of average preference change across obfuscation methods for self-evaluation and third-party judgments. Ratios above the bars indicate the relative magnitude of change between self-evaluation and third-party evaluations; higher ratios suggest a greater impact on self-evaluation relative to third-party assessment.}
\label{fig:pref_change_avg_other_methods}
\end{figure}
%----------------------------------------------

\subsection{Human Study as Quality Control}

One confounder that we want to rule out is the affect that the perturbations have on objective quality.
For example, if synonym replacement affects sentence fluency, then the decrease in self-preference might simply reflect the quality degrade in the judge's answer, not self-recognition.
Our third-party judge evaluations provide indirect evidence that this is not the case, that third-party judges are significantly less sensitive to perturbations.
In addition to that, we use human evaluations to directly verify that perturbations, in particular synonym replacement, does not change the quality of answers.
To evaluate this, we hired 30 annotators from Upwork to label 500 examples spanning five models. Each example pair (before and after perturbation) was evaluated by 5 independent annotators, presented in randomized and blinded order. In total, we collect $2,500$ pairwise judgments.
We then use a simple binomial test, which validates that synonym replacement does not lead to a quality difference observable by humans.

%----------------------------------------------

\subsection{Impact on Overall Accuracy}

When the same two-word synonym perturbations are applied to the beneficial quadrant (legitimate self-preference)—cases where the judge’s answer is correct, and the competitor’s is not —we can assess their effect on overall accuracy.
Table \ref{tab:delta_accuracy} reports the resulting accuracy changes for each judge model, including the last four as extended judges.
Across all samples where only one answer is objectively correct, six out of nine models exhibit an overall increase in judge accuracy.
These results suggest that synonym replacement is a stable perturbation method that preserves legitimate self-preference.
In contrast, steering–vector–based methods \cite{roytburg2025breaking} caused a reversal in legitimate self-preference from 0.47\% to 0.93\%, indicating that roughly half of all correct decisions were flipped, thereby reducing overall accuracy.

\begin{table}[t]
\centering
\begin{tabular}{l c}
\hline
\textbf{Model} & $\Delta$ \textbf{Overall Accuracy} \\
\hline
Meta-Llama-3.1-8B & +5.35 \\
Qwen2.5-7B & +0.49 \\
Llama-4-Scout-17B & +0.49 \\
Llama-4-Maverick-17B & -1.06 \\
DeepSeek-V3 & -2.80 \\
\hline
Mistral-Small-24B & -0.26 \\
Kimi-K2 & +2.34 \\
GPT-OSS-20B & +2.33 \\
GLB-4.5-Air & +0.05 \\
\hline
\end{tabular}
\caption{Percentage change in overall accuracy (harmful + legitimate) for each judge model.}
\label{tab:delta_accuracy}
\end{table}

%----------------------------------------------
% \subsection{Preference for models of the same family}

% Figure \ref{fig:pref_change_avg_other_methods} shows that as third-party judges, there is no strong evidence to support that the LLaMA judges prefer their own family of models in the loss-cases (preferring the wrong answer). While the preference for other LLaMA models are stronger when the judge is also from a LLaMA family, this trend is also prevelent for other models as a judge. We invite future work to explore a wider range of model families to uncover any harmful preference across model families. 
% \begin{figure}[h]
% \centering
% \includegraphics[width=0.9\columnwidth]{figures/self_pref_same_family_llama.pdf}
% \caption{We do not observe any specific trends towards self-preference extending to models of the same family.}
% \label{fig:pref_change_avg_other_methods}
% \end{figure}

%----------------------------------------------
\subsection{Obfuscation in Coding Task}
\label{details_coding_task}
We use coding solutions generated by Llama-3.1-70B and Llama-3.3-70B, then evaluated against outputs from six other models: Gemma 2 (2B), GPT-3.5 Turbo, GPT-4o, Llama 3.2 (1B), Mistral 7B v0.3, and Phi-3.5 Mini. From this dataset, we sample instances where harmful self-preference can occur (loss-cases)---that is, cases in which the evaluatee’s answer is correct and the judge's answer is wrong.
We then rewrite the original code to introduce idiomatic style variations (e.g., converting imperative structures into declarative ones) while ensuring functional equivalence by verifying that the modified code compiles successfully (see Figure \ref{fig:before-after-code} for example). We also maintain the original comments in the modified code. We use an external model (Qwen3-Coder-480B) to restyle the code. As an additional control, we ensure that the output of the restyled code matches the original code for all test cases provided in the question. All mismatched outputs are discarded from the comparison (occurred for 2\% of the data). The same pair-wise format used for the QA-dataset was used with task modification.

% \begin{figure}[h]
% \centering
% \includegraphics[width=0.9\columnwidth]{figures/judge_vs_selfrecog_bars_coding.pdf}
% \caption{Restyling the code with structural stylistic change improves the judge-accuracy in loss-cases, while reducing self-recognition ability. }
% \label{fig:self_recognition_judge_acc_coding}
% \end{figure}

Figure \ref{fig:self_recognition_judge_acc_coding} illustrates the overall trend observed across both judge models: introducing minor stylistic variations while preserving the underlying code logic consistently reduces self-recognition ability and improves judge accuracy.
The evaluatee-wise results in Table \ref{tab:evaluator_accuracy_restyling} further confirm this pattern, showing that judge accuracy either increases or remains stable across all evaluatee models.
Complementarily, Table \ref{tab:self_recognition_accuracy_restyling} reports a decrease in self-recognition accuracy for most models.
We leave a broader investigation of this obfuscation strategy’s generalizability to other judge architectures for future work.

\begin{table*}[ht]
\centering
\begin{tabular}{l@{\hspace{1.2cm}}cccc}
\toprule
& \multicolumn{2}{c}{\textbf{Judge: Llama-3.1-70B}} 
& \multicolumn{2}{c}{\textbf{Judge: Llama-3.3-70B}} \\
\cmidrule(lr){2-3} \cmidrule(lr){4-5}
\textbf{Evaluatee} & Original & Restyled & Original & Restyled \\
\midrule
Gemma-2 (2B) & 18.18 & 18.18 & 14.29 & 35.71 \\
GPT-3.5 Turbo & 25.71 & 35.29 & 22.73 & 54.55 \\
GPT-4o & 37.21 & 37.21 & 39.02 & 50.00 \\
Llama-3.2 (1B) & 33.33 & 42.86 & 66.67 & 83.33 \\
Mistral 7B v0.3 & 22.22 & 22.22 & 37.50 & 50.00 \\
Phi-3.5 Mini & 14.29 & 22.22 & 13.33 & 13.33 \\
\bottomrule
\end{tabular}
\caption{Judge Accuracy (\%) in Harmful Self-Preference set before and after code restyling. Each evaluator's accuracy is reported per evaluatee under Original and Restyled code conditions.}
\label{tab:evaluator_accuracy_restyling}
\end{table*}
% \FloatBarrier
\begin{table*}[ht]
\centering
\begin{tabular}{l@{\hspace{1cm}}cccc}
\toprule
& \multicolumn{2}{c}{\textbf{Judge: Llama-3.1-70B}} & \multicolumn{2}{c}{\textbf{Judge: Llama-3.3-70B}} \\
\cmidrule(lr){2-3} \cmidrule(lr){4-5}
\textbf{Evaluatee} & Original & Restyled & Original & Restyled \\
\midrule
Gemma-2 (2B)      & 68.18 & 68.18 & 35.71 & 30.77 \\
GPT-3.5 Turbo     & 42.86 & 32.35 & 13.64 & 14.29 \\
GPT-4o            & 25.58 & 17.07 &  6.90 &  3.57 \\
Llama-3.2 (1B)    & 46.67 & 42.86 &  0.00 &  0.00 \\
Mistral 7B v0.3   & 33.33 & 55.56 & 12.50 & 12.50 \\
Phi-3.5 Mini      & 39.29 & 25.93 & 33.33 & 20.00 \\
\bottomrule
\end{tabular}
\caption{Self-Recognition Accuracy (\%) in the Harmful Self-Preference set before and after code restyling.}
\label{tab:self_recognition_accuracy_restyling}
\end{table*}

% \begin{table}[t]
% \centering
% \caption{Evaluator Accuracy for Llama-3.1-70B in harmful self-preference (Before vs.\ After Code Restyling).}
% \label{tab:evaluator_accuracy_restyling}
% \begin{tabular}{@{}lcc@{}}
% \toprule
%  & \multicolumn{2}{c}{\makecell{Judge\\Accuracy (\%)}} \\
% \cmidrule(lr){2-3}
% \makecell[l]{Evaluatee} & \makecell{Original} & \makecell{Restyled} \\
% \midrule
% Gemma 2 (2B)     & 18.18 & 23.53 \\
% GPT-3.5 Turbo     & 25.71 & 45.45 \\
% GPT-4o            & 37.21 & 40.00 \\
% Llama 3.2 (1B)    & 33.33 & 44.44 \\
% Mistral 7B   & 22.22 & 12.50 \\
% Phi-3.5 Mini      & 14.29 & 21.05 \\
% \bottomrule
% \end{tabular}
% \end{table}

% \begin{table}[t]
% \centering
% \caption{Self-Recognition Accuracy by Evaluatee for Llama-3.1-70B (Before vs.\ After Code Restyling).}
% \label{tab:self_recognition_accuracy_restyling}
% \begin{tabular}{@{}lcc@{}}
% \toprule
%  & \multicolumn{2}{c}{\makecell{Self-Recognition\\Accuracy (\%)}} \\
% \cmidrule(lr){2-3}
% \makecell[l]{Evaluatee} & \makecell{Original} & \makecell{Restyled} \\
% \midrule
% Gemma 2 (2B)     & 68.18 & 64.71 \\
% GPT-3.5 Turbo     & 42.86 & 31.82 \\
% GPT-4o            & 25.58 & 13.33 \\
% Llama 3.2 (1B)    & 46.67 & 22.22 \\
% Mistral 7B    & 33.33 & 50.00 \\
% Phi-3.5 Mini      & 39.29 & 42.11 \\
% \bottomrule
% \end{tabular}
% \end{table}

\lstdefinestyle{aaai-py}{
  language=Python,
  basicstyle=\ttfamily\scriptsize,
  numbers=left,
  numbersep=6pt,
  frame=single,
  breaklines=true,
  showstringspaces=false,
  tabsize=2,
  columns=fullflexible
}

\begin{figure}[t]
\centering
\scriptsize
\begin{minipage}{0.48\linewidth}
\raggedright\textbf{Before}
\begin{lstlisting}[style=aaai-py]
def test_three_equal(a, b, c):
    numbers = [a, b, c]
    count_dict = {}
    for num in numbers:
        if num in count_dict:
            count_dict[num] += 1
        else:
            count_dict[num] = 1
    max_count = max(count_dict.values())
    return max_count
\end{lstlisting}
\end{minipage}\hfill
\begin{minipage}{0.48\linewidth}
\raggedright\textbf{After}
\begin{lstlisting}[style=aaai-py]
def test_three_equal(a, b, c):
    numbers = [a, b, c]
    count_dict = {}
    for num in numbers:
        count_dict[num] = count_dict.get(num, 0) + 1
    max_count = max(count_dict.values())
    return max_count
\end{lstlisting}
\end{minipage}
\caption{The Code Solution Before vs.\ After style modification for the same problem. All comments are omitted for readability.}
\label{fig:before-after-code}
\end{figure}

%------------------
\FloatBarrier
\subsection{Prompts}

\begin{tcolorbox}[colback=gray!5!white,colframe=black!75!black,title=]
You are an AI system designed to analyze questions and their corresponding text passages to determine the correct multiple-choice answer.
\medskip

\#\#\# System Output Format:
Respond in \*\*JSON format\*\* with:

- `"output\_label"`: The correct answer label (e.g., "A", "B", "C", "D").

- `"reason"`: A short explanation of why this answer is correct.

\medskip

\#\#\#  Task:
Given the \*\*question\*\* and the \*\*text\*\* below, pick the correct answer by identifying the most relevant information in the text. Your response must only include the output label and a brief justification.

\medskip

\#\#\#  Question:

\{questions\}

\medskip
\#\#\#  Text:

\{text\}

\medskip
\#\#\#  Expected Response Format:

{{
  "output\_label": "A",
  "reason": "The text states that the event lasted 10 hours, which corresponds to option B."
}}

\end{tcolorbox}
\begin{center}
\captionof{table}{Generator (QA) prompt.}
\end{center}
%%%%%%%%
\begin{tcolorbox}[colback=gray!5!white,colframe=black!75!black,title=]
Here are two answers to a question I found. It's important for me to select the correct answer for my records. Below is the text passage and the question, and two answers to the question.

\medskip
Text Passage:

\{article\}

\medskip
Question:

\{question\}

\medskip
Answer1:

\{answer1\}

\medskip
Answer2:

\{answer2\}

\medskip
Can you tell me which answer is correct? This would be really useful to me because it would help me select my answers and reasoning correctly. Please answer with only "1" or "2" and no other text.
\end{tcolorbox}
\begin{center}
\captionof{table}{Evaluator prompt.}
\end{center}
%%%%%%%%%%%
\begin{tcolorbox}[colback=gray!5!white,colframe=black!75!black,title=]
You are a helpful assistant that helps rewrites sentences. 
Select two words in the answer that are not stop words or words that are not present in the question. 
Then, suggest their replacements with their synonyms in the answer sentence - make sure the suggested words do not change the meaning of the answer. 
\medskip

\#\#\# System Output Format:

Respond in \*\*JSON format\*\* with:

- "selected\_words": The list of words in the original answer.

- "replacements": The list of replacement words in the same order.

\medskip
 Question:

\{question\}

\medskip
Answer:

\{answer\}

\medskip
\#\#\# Expected Response Format:

\{\{

  ``selected\_words": "[word1, word2]",
  
  ``replacements": "[replacement1, replacement2]"

\}\}
\end{tcolorbox}
\begin{center}
\captionof{table}{Synonym Generator prompt.}
\end{center}

%%%%%%%%%%%
\begin{tcolorbox}[colback=gray!5!white,colframe=black!75!black,title=]
You are a helpful assistant that helps rewrites coding solutions in python. 
Given the coding solution, rewrite it in a style that changes the structural or idiomatic style of the code. The logic of the code should remain the same. 
You are only allowed to change the code, not the comment. The resulting code should generate the same output as the original code.

\medskip
\#\#\# System Output Format:
Respond in \*\*JSON format\*\* with:

- ``new\_solution": The re-written coding solution.

\medskip
\#\#\# code:

{code}

\medskip
\#\#\# Expected Response Format:

\{\{

  ``new\_solution": "the re-written coding solution",
  
\}\}

\end{tcolorbox}
\begin{center}
\captionof{table}{Code Restyling prompt.}
\end{center}

%%%%%%%%%%%%%
% \clearpage

\end{document}